\documentclass[10pt,twocolumn,letterpaper]{article}

\usepackage{cvpr}              

\usepackage{graphicx}
\usepackage{wrapfig}
\usepackage{amsmath}
\usepackage{amssymb}
\usepackage{booktabs}
\usepackage{multirow}
\usepackage{colortbl}
\usepackage{xcolor}
\usepackage{multirow}
\usepackage{subcaption}
\usepackage{graphicx,wrapfig}
\usepackage{booktabs}
\usepackage{siunitx}
\definecolor{lightgray}{gray}{0.9}
\definecolor{linecolor}{rgb}{0.82, 0.94, 0.75}
\usepackage{siunitx}
\definecolor{evaunit01green}{RGB}{82,208,83}
\definecolor{lowred}{RGB}{238,18,137}

\definecolor{lowerred}{RGB}{255,110,180}
\definecolor{bgreen}{RGB}{0,170,0}
\definecolor{bred}{RGB}{220,0,0}
\definecolor{mydarkblue}{RGB}{0,0,150}
\definecolor{Gray}{gray}{0.93}
\definecolor{skyblue}{rgb}{0.925,0.957,1}
\definecolor{icmlblue}{rgb}{0,0.08,0.45} 
\definecolor{PineGreen}{RGB}{1, 121, 111}
\definecolor{RedBrick}{RGB}{77,0,38}
\definecolor{linecolor3}{HTML}{E3EFF7}
\definecolor{linecolor4}{HTML}{F1F7FB}

\usepackage{enumitem}

\usepackage[pagebackref,breaklinks,colorlinks]{hyperref}

\usepackage{bbm}
\usepackage[capitalize]{cleveref}
\crefname{section}{Sec.}{Secs.}
\Crefname{section}{Section}{Sections}
\Crefname{table}{Table}{Tables}
\crefname{table}{Tab.}{Tabs.}

\def\confName{CVPR}
\def\confYear{2026}

\usepackage{tablefootnote}
\begin{document}

\title{Point Cloud as a Foreign Language for Multi-modal Large Language Model}

\author{Sneha Paul, Zachary Patterson, Nizar Bouguila \\
Concordia University,
Canada\\
{\tt\small sneha.paul@mail.concordia.ca},
{\tt\small \{zachary.patterson, nizar.bouguila\}@concordia.ca}
}
\maketitle

\begin{abstract}
Multi-modal large language models (MLLMs) have shown remarkable progress in integrating visual and linguistic understanding. Recent efforts have extended these capabilities to 3D understanding through encoder-based architectures that rely on pre-trained 3D encoders to extract geometric features. However, such approaches suffer from semantic misalignment between geometric and linguistic spaces, resolution sensitivity, and substantial computational overhead.
In this work, we present SAGE, the first end-to-end 3D MLLM that directly processes raw point clouds without relying on a pre-trained 3D encoder. Our approach introduces a lightweight 3D tokenizer that combines geometric sampling and neighbourhood aggregation with vector quantization to convert point clouds into discrete tokens—treating 3D data as a foreign language that naturally extends the LLM’s vocabulary.
Furthermore, to enhance the model’s reasoning capability on complex 3D tasks, we propose a preference optimization training strategy with a semantic alignment–based reward, specifically designed for open-ended 3D question answering where responses are descriptive. Extensive experiments across diverse 3D understanding benchmarks demonstrate that our end-to-end approach outperforms existing encoder-based methods while offering significant advantages in computational efficiency, generalization across LLM backbones, and robustness to input resolution variations. Code is available at:\href{https://github.com/snehaputul/SAGE3D}
{https://github.com/snehaputul/SAGE3D}.

\end{abstract}

\section{Introduction}
\label{sec:intro}
\begin{figure}
    \centering
    \includegraphics[width=\linewidth]{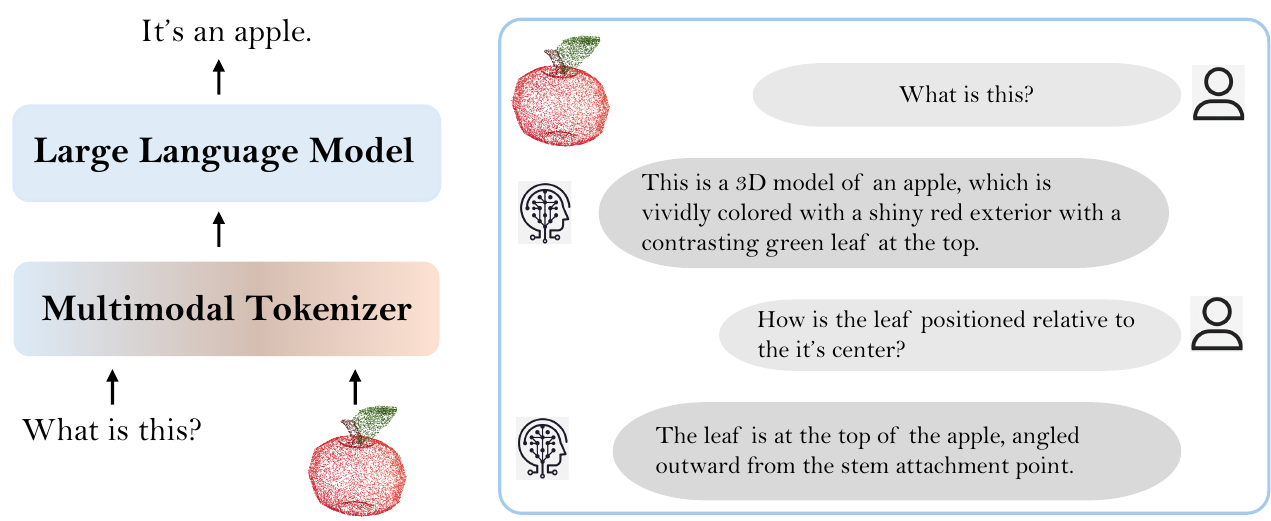}%
    \caption{Our proposed encoder-free 3D Multimodal Large Language Model efficiently captures 3D information from point clouds without relying on any pretrained 3D encoder. The figure on the left illustrates the overall architecture, while the figure on the right shows an example conversation about an object generated by our model. }
    \label{fig:banner}
    \vspace{-10pt}
\end{figure}
Large Language Models (LLMs)~\cite{llama, qwen, mistral, phi} have demonstrated remarkable capabilities in understanding complex linguistic tasks and solving challenging problems, including multi-step reasoning and advanced question answering. Inspired by their success in language domains, recent research has extended the capabilities of LLMs toward multimodal understanding, leading to the development of Multi-modal large language models (MLLM)~\cite{llavanext, gpt4v}. These models integrate multiple modalities—such as visual and auditory signals—with language, enabling unified and cross-modal reasoning. Building on this progress, recent research \cite{pointllm} has explored extending MLLMs with 3D understanding capabilities, enabling models to reason about and generate content grounded in spatial structures by incorporating point cloud data into the multimodal understanding framework. By coupling 3D spatial understanding with the reasoning power of LLMs, 3D MLLMs hold the promise of advancing high-level perception, embodied intelligence, and human–robot interaction through natural 3D–language reasoning and generation.

Current 3D-MLLMs primarily follow a pre-trained encoder-dependent design paradigm, where a pre-trained 3D encoder extracts geometric embeddings that are projected into the LLM’s input space. While this design benefits from the representational capacity of pre-trained 3D encoders~\cite{pointbert, zhang2023i2pmae, pang2022masked3d, tang2024geometryclip}, it faces few key limitations (which may not necessarily be unique to 3D but constitute the focus of our work): (1) Semantic misalignment: Existing pre-trained 3D encoders are generally trained with self-supervised or contrastive loss~\cite{xie2020pointcontrast, qi2023contrastive3d, pang2022masked3d} optimized for geometric discrimination rather than linguistic grounding, resulting in embeddings that remain semantically incompatible with the LLM’s input space. Yet, the assumption of the availability of a large pre-trained encoder makes it impractical for domains with limited in-distribution data, where pre-training such encoders is infeasible.
(2) Resolution mismatch: Existing 3D encoders assume fixed input sizes (e.g., 8,192 points in Point-BERT), whereas real-world point clouds vary widely in density, leading to degraded representations when dense point clouds lose fine-grained details through downsampling, or when sparse ones require upsampling that introduces geometric artifacts or incomplete coverage. (3) Computational overhead: A large 3D encoder introduces considerable processing overhead before the LLM can begin generating responses, slowing down inference and increasing resource demands.

To address the aforementioned limitations, we explore an end-to-end design for 3D MLLMs that does not rely on a pre-trained encoder, but instead learns 3D representations jointly with the LLM through a 3D tokenizer. 
It is important to clarify that “encoder-free” does not imply a parameter-free approach---the tokenizer still contains a small number of learnable parameters, but is significantly smaller than those of a typical encoder. Moreover, techniques effective in related domains (e.g., vision)—such as dividing images into patches and projecting them directly into the language model’s input space—prove ineffective for 3D data, as they fail to capture the underlying geometric structure and relative spatial relationships fundamental to 3D understanding.

To this end, we introduce SAGE---a Spatial-Aware GEnerative model that eliminates the need for pre-trained 3D encoders and enables direct 3D-to-language generation. Central to our approach is a novel lightweight 3D tokenizer that projects raw 3D inputs into the LLM’s input space, effectively allowing the model to interpret 3D data as a foreign language. Specifically, we employ farthest point sampling and nearest-neighbour grouping to capture local geometric structures, yielding spatially-aware representations that encode both positional and semantic cues of 3D input. To bridge the gap between continuous geometric features and the discrete token space of the LLM, we apply vector quantization with a learnable codebook that discretizes 3D features into a finite vocabulary of tokens—serving as an extension of the LLM tokenizer for 3D understanding.

To further enhance the reasoning capability of our 3D MLLM, we introduce an RL-based preference optimization training strategy aimed at improving performance on challenging 3D understanding tasks that require complex reasoning. Since most existing RL-based techniques, such as GRPO, rely on rewards derived from the correctness of final answers on verifiable tasks (e.g., mathematical reasoning), they cannot be directly applied to 3D question-answering tasks, where answers are often descriptive and not objectively verifiable. To address this limitation, we propose a novel reward formulation based on semantic alignment between the generated response and the ground-truth answer, enabling effective RL-based optimization for open-ended 3D reasoning. While the preference optimization stage is designed to enhance complex 3D understanding, we report results for both the complete model with preference optimization (SAGE) and the variant without it (SAGE$^*$)—the proposed encoder-free end-to-end 3D MLLM trained under the standard protocol—to validate the effectiveness of our model.

For empirical evaluation, we conduct extensive experiments across standard 3D understanding benchmarks, including captioning, question answering, and recognition. Our results yield two key findings:
(1) SAGE$^*$, despite having no pre-trained encoder, matches or surpasses the performance of existing 3D MLLMs that rely on extensive pre-training. Additionally, it exhibits significantly higher inference efficiency, robustness to input resolution changes. We report this variant as SAGE$^*$, and report all the main results separately for this.
(2) SAGE, trained with the proposed preference optimization, further outperforms all existing methods across multiple benchmarks, with particularly strong gains on complex understanding tasks. 
Overall, we make the following contributions in this paper:
\begin{itemize}
    \item We introduce SAGE, the first end-to-end 3D MLLM that directly processes raw point clouds without relying on any pre-trained 3D encoders. Our lightweight 3D tokenizer integrates geometric sampling and neighbourhood aggregation with vector quantization to convert raw 3D data into discrete tokens—treating point clouds as a foreign language that naturally extends the LLM’s vocabulary while preserving geometric structure and positional cues.
 
    \item We introduce a preference optimization strategy to enhance complex reasoning in 3D MLLMs. As existing RL methods rely on verifiable rewards and cannot handle descriptive 3D tasks, we propose a semantic alignment–based reward that enables effective reinforcement learning for open-ended 3D reasoning.
    
    \item Extensive experiments show that SAGE$^*$ matches or exceeds pre-trained encoder–based methods while being more efficient and robust, and SAGE with preference optimization outperforms existing methods.
\end{itemize}

\section{Related Works}
\label{sec:literature_review}

\subsection{3D MLLM}

Building on multimodal architectures from related domains, recent work has significantly advanced 3D vision-language modelling. Early approaches such as PointLLM \cite{pointbind} and ShapeLLM \cite{shapellm} established the encoder-based paradigm by coupling pre-trained 3D encoders with projection modules to bridge geometric and linguistic representations. While effective, this design inherits fundamental limitations from the assumed availability of large-scale pre-trained encoders. Specifically, methods like LLaVA-3D \cite{llava3d}, Robin3D \cite{robin3d}, and 3D-LLaVA \cite{3dllava} extend instruction-following capabilities into 3D by incorporating specialized projectors or spatial priors, yet remain constrained by the semantic misalignment between encoder outputs—typically trained with self-supervised objectives \cite{pang2023masked} or contrastive losses \cite{xie2020pointcontrast}—and the linguistic space of LLMs. This misalignment necessitates complex projection architectures that struggle to reconcile geometric discrimination with linguistic grounding.

A complementary line of work explores reconstructive and instance-aware objectives to improve grounding. Ross3D \cite{ross3d} and Inst3D-LMM \cite{inst3d} incorporate auxiliary losses for relational reasoning, while methods like SpatialPIN \cite{spatialpin}, MSR3D \cite{msr3d}, and Situation3D \cite{situational} demonstrate that explicit geometric priors enhance spatial relation understanding. However, these approaches compound architectural complexity through handcrafted geometry-aware modules, and the fixed-resolution assumption of pre-trained encoders (e.g., 8,192 points in Point-BERT \cite{pointbert}) creates brittleness when handling variable-density point clouds—dense inputs lose fine-grained details through downsampling, while sparse inputs require upsampling that introduces artifacts.

Temporal and memory-based extensions such as GPT4Scene \cite{gpt4scene}, Video-3D LLM \cite{video3d}, and 3DLLM-Mem \cite{3dllmmem} enhance reasoning across sequential observations but inherit the computational overhead of encoder preprocessing, making real-time deployment challenging. Similarly, grounding-focused methods like SeeGround \cite{seeground} and SplatTalk \cite{splattalk} achieve impressive zero-shot capabilities but remain dependent on encoder-extracted features. Efficiency-oriented approaches, including GreenPLM \cite{greenplm}, Point-Bind \cite{pointbind}, and MiniGPT-3D \cite{minigpt}, reduce annotation requirements through cross-modal alignment but do not address the fundamental encoder dependency that limits generalization across LLM backbones and input resolutions.

\subsection{Encoder-free MLLM}

The limitations of encoder-based designs have motivated recent exploration of encoder-free architectures in 2D vision-language domains. Fuyu-8B \cite{fuyu} and Otter-HD \cite{otter} pioneered decoder-only models that project raw image patches directly into language models, demonstrating architectural simplicity but achieving moderate performance due to insufficient geometric structure preservation. The EVE series \cite{eve, evev2} advanced this direction by unifying vision and language representations within a single decoder through auxiliary supervision and scaled training data. Similarly, SHOW-O \cite{showo} proposed unified understanding and generation without dedicated encoders. Alternative approaches employ VQ tokenizers \cite{chameleon} or lightweight projection layers \cite{eve, solo} to convert images into discrete token sequences, treating visual data as an extension of the language model's vocabulary.
However, methods successful in 2D vision---particularly patch-based tokenization---cannot be directly applied to 3D point clouds, which lack regular grid topology and necessitate explicit preservation of local geometric structure and inter-point spatial relationships that patch-based approaches inherently discard. ENEL \cite{enel} is a concurrent work in an attempt towards endoer-free learning, yet its tokenizer contains a large number of parameters.

\begin{figure}
    \centering
    \includegraphics[width=\linewidth]{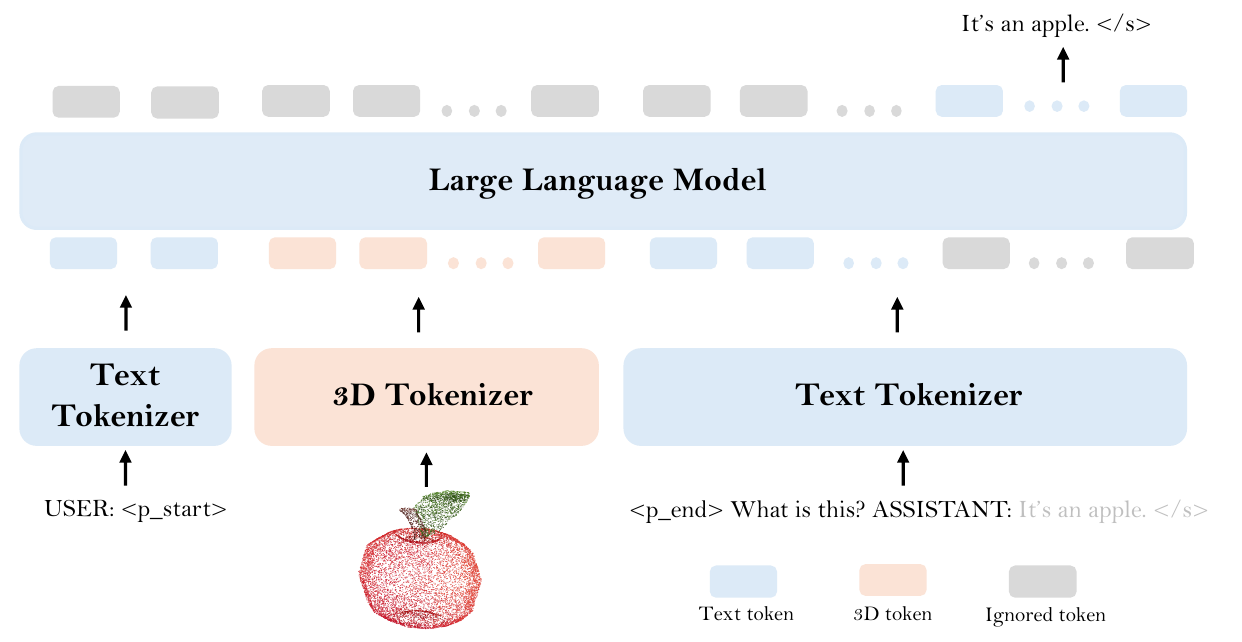}%
    \caption{Architecture of our proposed method, encoder-free 3D Multimodal Large Language Model.}
    \label{fig:method}
    \vspace{-10pt}
\end{figure}

\section{Method}

\subsection{Preliminaries}

Let $\mathcal{P} \in \mathbb{R}^{N \times 3}$ denote a point cloud and $q$ be a natural language question about its semantic or geometric properties. A 3D multimodal large language model such as PointLLM \cite{pointllm} aims to generate a coherent textual response $r$ conditioned on both $\mathcal{P}$ and $q$, which comprises three components: a pre-trained 3D encoder that extracts geometric features from raw point clouds, a projection module that maps these features into the LLM's embedding space, and a pre-trained large language model (LLM) backbone $f_{\text{llm}}$. Here, $f_{\text{llm}}$ is a decoder-only Transformer \cite{vaswani2017attention} that processes a heterogeneous sequence consisting of text and point cloud tokens. Let $\mathbf{T} = (t_1, t_2, \dots, t_K)$ denote this input sequence of $K$ discrete token IDs. Through self-attention, $f_{\text{llm}}$ captures dependencies across modalities, enabling joint reasoning over textual and 3D geometric inputs.

Each token $t_i$ is embedded into $\mathbb{R}^{d_{\text{llm}}}$ and the LLM autoregressively produces hidden states $\hat{\mathbf{h}}_i = f_{\text{llm}}(\mathbf{T}_{<i})$, where $\mathbf{T}_{<i} = (t_1, \dots, t_{i-1})$. Each $\hat{\mathbf{h}}_i$ is projected to vocabulary logits via $f_{\text{vocab}} : \mathbb{R}^{d_{\text{llm}}} \to \mathbb{R}^{|\mathcal{V}|}$, and the next token is predicted as $\tilde{t}_i = \arg\max_{w \in \mathcal{V}} f_{\text{vocab}}(\hat{\mathbf{h}}_i)[w]$. Training minimizes negative log-likelihood over response tokens only, masking out instruction tokens. As mentioned earlier, such an encoder-based approach has several drawbacks, including semantic misalignment, resolution mismatch, and computational overhead. To this end, we explore an end-to-end 3D MLLM approach that is pre-trained-encoder free.

\subsection{End-to-end 3D MLLM}
The core idea of our proposed approach (Figure \ref{fig:method}) is to treat a point cloud as a foreign language without using a pre-trained 3D encoder. To extend the vocabulary of pre-trained LLM tokenizer to understand point clouds, we introduce a lightweight, trainable tokenizer that projects the point cloud into the LLM's input space. Given an input point cloud $\mathcal{P} \in \mathbb{R}^{N \times D}$, where $N$ denotes the number of points and $D$ the dimensionality of each point feature (e.g., $D=6$ for $(x, y, z, r, g, b)$ when including RGB color information), the proposed tokenizer outputs a sequence of quantized tokens $\mathbf{Y} = (\mathbf{y}_1, \mathbf{y}_2, \dots, \mathbf{y}_M)$, each selected from a discrete learned codebook, with $M$ denoting the number of point tokens. Below, we discuss the main components of the proposed tokenizer.

\textbf{Geometric Sampling and Grouping.}
Given the dense point cloud $\mathcal{P}$, we first sample a representative set of points $N_s$ using Farthest Point Sampling (FPS). These points represent cluster centres for each of which $K_g$-nearest neighbours are identified using the KNN algorithm to form local sub-clouds that preserve neighbourhood geometry. Next, a local geometry aggregation module constructs spatially contextualized representations that capture both geometric and semantic cues by projecting the point feature into geometric feature space, adding relative positional embedding, and applying global max-pooling to each sub-cloud. Specifically, aggregated features are pooled to obtain the compact latent representation $\mathbf{Z} \in \mathbb{R}^{M \times d_g}$, where $d_g$ denotes the geometric feature dimension and $M$ is the number of spatial tokens retained after pooling. 

\textbf{Projection to LLM Space.}
These features are first projected into the language model embedding space using a learnable projection matrix $\mathbf{W} \in \mathbb{R}^{d_g \times d_{\text{llm}}}$:
\begin{equation}
\mathbf{H} = \mathbf{Z} \mathbf{W} \in \mathbb{R}^{M \times d_{\text{llm}}}.
\end{equation}

\textbf{Vector Quantization.}
To bridge the gap between continuous geometric features and discrete language tokens, we employ vector quantization with a learnable codebook that maps continuous embeddings into a finite vocabulary of 3D tokens. This process effectively extends the LLM tokenizer to the 3D domain, treating geometric representations as a new language.
To discretize the projected features $\mathbf{H}$, we define a learnable codebook $\mathcal{C} = \{\mathbf{e}_k\}_{k=1}^{C}$, where $C$ denotes the codebook size and each $\mathbf{e}_k \in \mathbb{R}^{d_{\text{llm}}}$ represents a code vector. Each latent feature $\mathbf{h}_i$ (the $i$-th row of $\mathbf{H}$) is quantized by selecting its nearest codebook entry:
\begin{equation}
    q(\mathbf{h}_i) = \arg\min_{k} \|\mathbf{h}_i - \mathbf{e}_k\|_2^2, 
    \quad 
    \mathbf{H}_q = \{\mathbf{e}_{q(\mathbf{h}_i)}\}_{i=1}^{M},
\end{equation}
where $\mathbf{H}_q \in \mathbb{R}^{M \times d_{\text{llm}}}$ denotes the quantized representation. The codebook $\mathcal{C}$ is trained jointly with the rest of the model, allowing the discovery of discrete 3D primitives that align with both geometric and linguistic semantics.

Since the quantization operation is inherently non-differentiable, we employ the vector quantization loss \cite{van2017neural} to enable gradient-based training:
\begin{equation}
\mathcal{L}_{\text{VQ}} = 
\underbrace{\|\text{sg}[\mathbf{H}] - \mathbf{H}_q\|_2^2}_{\text{codebook loss}} 
+ 
\beta \underbrace{\|\mathbf{H} - \text{sg}[\mathbf{H}_q]\|_2^2}_{\text{commitment loss}},
\end{equation}
where $\text{sg}[\cdot]$ denotes the stop-gradient operator. The first term updates the codebook vectors to better represent the projected features, while the second term constrains the projection to produce features close to existing codebook entries, preventing drift from the quantized space.

\textbf{Mixed-Modality Sequence Formation.}
After vector quantization, the discretized tokens from the codebook are directly concatenated with text tokens to form a mixed-modality input sequence for the LLM:
\begin{equation}
[\texttt{<p\_start>}, \mathbf{e}_{q(\mathbf{h}_1)}, \dots, \mathbf{e}_{q(\mathbf{h}_M)}, \texttt{<p\_end>}, \mathbf{w}_1, \dots, \mathbf{w}_L],
\end{equation}
where $\{\mathbf{w}_j\}_{j=1}^{L}$ are text token embeddings and $\{\mathbf{e}_{q(\mathbf{h}_i)}\}_{i=1}^{M}$ are the quantized 3D token embeddings retrieved from the codebook, all in $\mathbb{R}^{d_{\text{llm}}}$. \texttt{<p\_start>} and \texttt{<p\_end>} are special tokens indicating the start and end of point cloud tokens. The entire framework is optimized end-to-end using a combined objective:
\begin{equation}
    \mathcal{L}_{\text{total}} = \mathcal{L}_{\text{NTP}} + \lambda \mathcal{L}_{\text{VQ}},
\end{equation}
where $\mathcal{L}_{\text{NTP}}$ is the next-token prediction loss, and $\lambda$ balances the two terms.

\subsection{Proposed Training Pipeline}
\begin{figure}
    \centering
    \includegraphics[width=\linewidth]{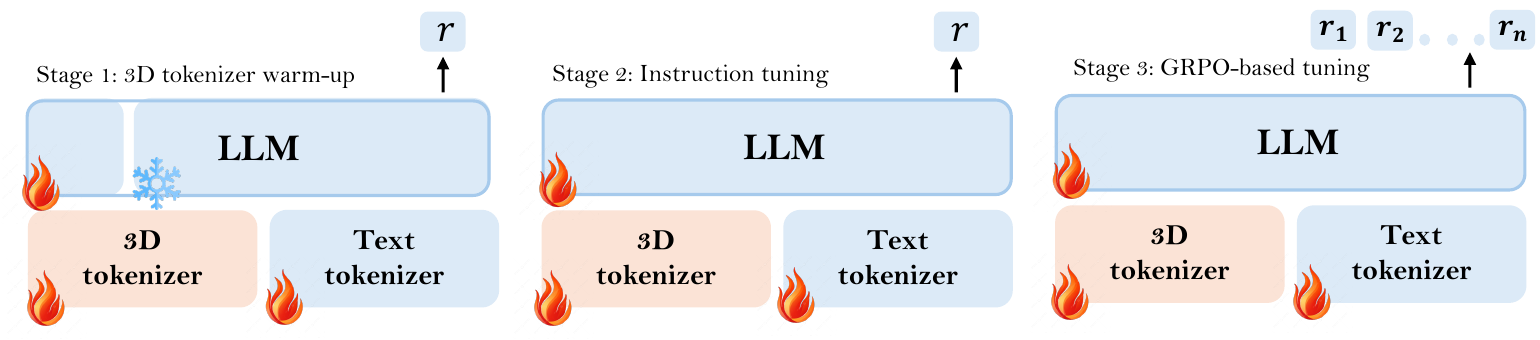}%
    \caption{The proposed training pipeline of our model. The model is trained in three stages --- each stage focusing on a specific training objective. }
    \label{fig:pipeline}
    \vspace{-10pt}
\end{figure}

While our end-to-end 3D MLLM can be seamlessly integrated into existing multimodal training frameworks~\cite{pointllm, shapellm}, and our empirical results in the section \ref{sec:experiments} show improved performance over existing methods, we introduce a preference optimization strategy to further improve the reasoning capability of 3D MLLM for more complex tasks. This stage is employed on top of the existing two-stage training protocol adopted by the existing literature. Following, we describe the whole pipeline in detail:

\textbf{Stage 1: 3D Tokenizer Warm-up.}  
In the first stage, the newly initialized 3D tokenizer is jointly trained with a small subset of the LLM layers using a next-token prediction objective on 3D captioning datasets. 
Specialized structural tokens, $\texttt{<p\_start>}$ and $\texttt{<p\_end>}$, are introduced to delineate 3D token sequences within the language stream. 
This stage focuses on aligning geometric token embeddings with the linguistic representation space while stabilizing early-stage multimodal interactions.

\textbf{Stage 2: Instruction Tuning.}  
The second stage performs end-to-end instruction tuning using multimodal instruction–response pairs. 
Here, the entire model (tokenizer and LLM) is optimized to comprehend and generate contextually accurate outputs conditioned jointly on textual and 3D point cloud inputs. 
This step substantially enhances the model's cross-modal reasoning and alignment with instruction-following behaviours.

\textbf{Stage 3: Preference Optimization.}  
Finally, we introduce a new preference optimization stage that utilizes a novel reward function with Group Relative Policy Optimization (GRPO)~\cite{grpo} to fine-tune the model toward preference-aligned responses. 
Unlike most common reinforcement learning approaches, such as PPO~\cite{schulman2017proximal}, GRPO eliminates the need for an explicit reward model by utilizing the model's own likelihoods to derive relative quality preferences across a set of generated responses, leading to more stable and efficient optimization.

While GRPO has been primarily adopted for verifiable domains such as mathematical reasoning—where correctness can be explicitly determined by comparing the final answer to a ground truth—3D understanding tasks are inherently descriptive and often admit multiple valid responses expressed in different wordings. 
To adapt GRPO to this open-ended setting, we design a reward function that combines semantic similarity with length regularization to evaluate generated responses in a continuous and interpretable manner. 
For each instruction–point cloud pair $(q, \mathcal{P})$, the model generates a group of $m$ candidate responses $\{y_1, y_2, \ldots, y_m\}$, each compared against a reference response $y_{\text{ref}}$. 
We employ a pre-trained sentence encoder (Sentence-BERT) $\mathcal{E}(\cdot)$ to encode both the generated and reference sentences, and compute their cosine similarity:
\begin{equation}
s^{(\text{sem})}_i = 
\frac{\mathcal{E}(y_i) \cdot \mathcal{E}(y_{\text{ref}})}
{\|\mathcal{E}(y_i)\|_2 \, \|\mathcal{E}(y_{\text{ref}})\|_2},
\quad i = 1, 2, \ldots, m.
\end{equation}
This term provides a continuous measure of semantic alignment, rewarding responses that are meaningfully close to the reference description, even if phrased differently.

To further discourage excessively short or overly long outputs, we introduce a length-based reward defined as:
\begin{equation}
s^{(\text{len})}_i = \exp\!\left(-\frac{(L_i - L_{\text{ref}})^2}{2\sigma^2}\right),
\end{equation}
where $L_i$ and $L_{\text{ref}}$ denote the token lengths of the generated and reference responses, respectively, and $\sigma$ controls the tolerance around the reference length. 
This reward reaches its maximum when $L_i = L_{\text{ref}}$ and decays smoothly as the length deviation increases.

The final outcome score for each response is a weighted combination of both terms:
\begin{equation}
s_i = \alpha\, s^{(\text{sem})}_i + (1 - \alpha)\, s^{(\text{len})}_i,
\end{equation}
where $\alpha \in [0,1]$ balances semantic accuracy and length regularization. 
This composite reward enables stable GRPO optimization in descriptive 3D–language tasks, ensuring the model learns to produce semantically relevant and appropriately concise responses.

Given the outcome scores $\{s_i\}_{i=1}^m$ computed from our composite reward, we compute a normalized group-wise advantage for each sampled response:
\begin{equation}
A_i = 
\frac{s_i - \bar{s}}
{\sqrt{\frac{1}{m} \sum_{j=1}^{m} (s_j - \bar{s})^2 + \epsilon}},
\quad
\text{where} \quad
\bar{s} = \frac{1}{m} \sum_{j=1}^{m} s_j.
\end{equation}
Here, $\epsilon$ is a small constant ($1\times10^{-9}$) added for numerical stability to prevent division by zero. This normalization ensures that the relative quality of responses, rather than their absolute magnitude, drives learning within each sampled group. 
The GRPO objective is then defined as:
\begin{equation}
\mathcal{L}_{\text{GRPO}}(\theta)
= - \frac{1}{m} \sum_{i=1}^{m} A_i \log \pi_\theta(y_i \mid q, \mathcal{P}),
\end{equation}
where $\pi_\theta(y_i \mid q, \mathcal{P})$ denotes the model's probability of generating response $y_i$ given the instruction–point cloud pair $(q, \mathcal{P})$.
This objective encourages the model to increase the likelihood of responses with higher relative rewards within each group, promoting self-improvement through outcome-based preference optimization.
During this stage, we reuse the instruction–response pairs from Stage~2, but sample multiple candidate responses per instruction and apply GRPO updates iteratively.

\section{Experiments}
\label{sec:experiments}

\subsection{Implementations Details}
\textbf{Training Details.}
Following the experimental setup of existing 3D–language alignment methods and to ensure fair comparison with prior work, we adopt {LLaMA} \cite{llama} as the backbone language model and initialize from the {Vicuna-7B v1.1} \cite{chiang2023vicuna} checkpoint. All implementation details for the first two stages follow those of PointLLM \cite{pointllm}, unless otherwise specified. 
All experiments are conducted on 4 NVIDIA H100 80GB GPUs in \texttt{BF16} precision.
We employ the \texttt{AdamW} optimizer with a cosine learning rate schedule, a warmup ratio of 0.03, and weight decay of 0.05. \texttt{FlashAttention} is used to improve memory efficiency during training.

During \textit{stage~1 (3D tokenizer warm-up),} the model is trained for 3 epochs with a batch size of 128 and an initial learning rate of $4\times10^{-4}$. During this stage, the 3D tokenizer module and the first four transformer layers of the LLM are jointly trained, while the remaining layers are frozen. Specialized point tokens \texttt{<p\_start>} and \texttt{<p\_end>} are randomly initialized and trained along with the model to denote point cloud boundaries.  
For the 3D tokenizer, we set the number of sampled points to $N_s=512$ and the number of neighbours to $K_g=81$. For vector quantization, we use a learnable codebook of size 8192. The weighting coefficient of vector quantization loss is $\beta = 0.25$ and the regularization coefficient in the total loss is $\lambda = 0.50$. 

\textit{In stage~2 (instruction tuning),} the model is fine-tuned end-to-end for 3 epochs with a batch size of 32 and a learning rate of $2\times10^{-5}$. Instruction–response pairs containing complex multi-modal queries are used to enhance the model's reasoning and alignment between text and 3D content.

Finally, in \textit{stage~3 (preference optimization),} the model is fully fine-tuned using the GRPO objective with a learning rate of $1\times10^{-6}$, and number of generated responses, $m=8$, trained for 1 epoch on the instruction tuning data. The weighting coefficient for group normalization is set to $\alpha=0.95$. The model-specific hyperparameters are listed in \textit{Appendix}~\ref{app:implementation} for reproducibility. Furthermore, an in-depth sensitivity analysis of these hyperparameters is also presented in \textit{Appendix}~\ref{app:ablation}.

We follow the evaluation protocol of prior works and report multiple metrics for 3D captioning, including {BLEU-1}, {ROUGE-L}, {METEOR}, and embedding-based scores using {SimCSE} and {Sentence-BERT}. For instruction-following and classification tasks, we additionally employ {GPT-4} as an external evaluator following \cite{pointllm, shapellm}, using Prompt 1 (\textit{Appendix} \ref{app:implementation}).

\textbf{Dataset Details.}
We train our model using the large-scale point–text instruction-following dataset introduced in {PointLLM}~\cite{pointllm}, which provides comprehensive coverage for 3D-language alignment. The dataset contains over {730K} point–text pairs across approximately {660K} unique 3D objects sourced from {Objaverse}~\cite{deitke2023objaverse}. It consists of {660K} brief caption-style instructions generated via Cap3D~\cite{luo2023scalable} and an additional {70K} complex instruction samples synthesized using {GPT-4}. The latter include {40K} single-turn conversations and {15K} multi-turn dialogues, designed to capture diverse object properties such as category, appearance, affordance, and function. For evaluation, we adopt the benchmark protocols established in PointLLM~\cite{pointllm}. Specifically, we evaluate on the same evaluation set as existing work (Objaverse~\cite{deitke2023objaverse}), each paired with high-quality Cap3D~\cite{luo2023scalable} captions. For 3D visual question answering (VQA) evaluation, we use the MM-Vet dataset \cite{shapellm}.

\subsection{Multimodal Understanding}
In this section, we discuss the main results of our proposed method and its comparison to existing methods.
\begin{table*}[!t]
\centering
\caption{Performance comparison on various 3D downstream tasks. Here, the Objaverse dataset is used for 3D object captioning and recognition tasks, and the MM-Vet dataset is used for 3D VQA. In addition to our full model \textit{(SAGE)}, we provide the results without the preference optimization \textit{(SAGE$^*$)}. Additional comparisons with existing methods on different datasets are provided in the \textit{Appendix} \ref{app:results}.}
\label{tab:main}
\small
\begin{tabular}{l|cccccc|c|c}
\hline
\multirow{2}{*}{\textbf{Model}} & \multicolumn{6}{c|}{\textbf{Captioning}} & {\textbf{Cls.}} & {\textbf{VQA}} \\
\cline{2-9}
& \textbf{GPT-4} & \textbf{Sentence-BERT} & \textbf{SimCSE} & \textbf{BLEU-1} & \textbf{ROUGE-L} & \textbf{METEOR} & \textbf{GPT-4} & \textbf{GPT-4} \\
\hline
InstructBLIP-7B \cite{dai2023instructblip} & 43.54 & 47.41 & 48.48 & 4.27 & 8.28 & 12.99 & 43.50 & -- \\
InstructBLIP-13B \cite{dai2023instructblip} & 44.97 & 45.90 & 48.86 & 4.65 & 8.85 & 13.23 & 44.25 & -- \\
LLaVA-7B \cite{liu2023visual} & 46.71 & 45.61 & 47.10 & 3.64 & 7.70 & 12.14 & 50.00 & -- \\
LLaVA-13B \cite{liu2023visual} & 38.28 & 46.37 & 45.90 & 4.02 & 8.15 & 12.58 & 51.75 & 47.90 \\
\hline
PointLLM-7B \cite{pointllm} & 44.85 & 47.47 & 48.55 & 3.87 & 7.30 & 11.92 & 53.00 & 41.20 \\
PointLLM-13B \cite{pointllm}  & 48.15 & 47.91 & 49.12 & 3.83 & 7.23 & 12.26 & 54.00 & 46.60 \\
ShapeLLM-7B \cite{shapellm} & 46.92 & 48.20 & 49.23 & -- & -- & -- & 54.50 & 47.40 \\
ShapeLLM-13B \cite{shapellm} & 48.94 & 48.52 & 49.98 & -- & -- & -- & 54.00 & {53.10} \\
\hline
\rowcolor{linecolor4}\textbf{SAGE-7B$^*$} & 49.05 & 49.23 & 48.56 & 7.41 & 10.25 & 14.35 & 55.71 &  46.38\\
\rowcolor{linecolor4}\textbf{SAGE-7B} & {50.98} & 50.11 & 49.70 & 9.50 & 12.66 & 16.95 & 57.11 &  49.53\\
\rowcolor{linecolor3}\textbf{SAGE-13B$^*$} & 48.54 & 48.99 & 50.18 & 7.98 & 12.48 & 14.27 & 56.39 & 53.21 \\
\rowcolor{linecolor3}\textbf{SAGE-13B} & {52.87} & {51.91} & {51.03} & 9.72 & 13.25 & 16.99 & {58.48} & 54.89 \\
\hline
\end{tabular}
\end{table*}

\textbf{3D Object Captioning.}
Table~\ref{tab:main} summarizes the quantitative results on the Objaverse 3D captioning benchmark following the evaluation protocol of PointLLM~\cite{pointllm}.  
Each model is prompted with “\textit{Caption this 3D model in detail}.”  
Our proposed model, SAGE, substantially outperforms all baselines across both GPT-4–based and text-similarity metrics.  
Specifically, our 7B and 13B variants achieve GPT-4 scores of {50.98} and {52.87}, surpassing ShapeLLM-13B by +3.93 and PointLLM-13B by +4.72, respectively.  
Our model also achieves a large gain in traditional lexical metrics—BLEU-1, ROUGE-L, and METEOR over PointLLM and SpaheLLM on both model sizes, suggesting that the learned 3D token representations contribute to richer and more precise linguistic grounding. Consistent improvements are observed in more advanced similarity-based measures such as Sentence-BERT and SimCSE, with +1.91 and +0.47 improvement for 7B and +3.93 and +1.05 for 13B.

\textbf{3D Object Classification.}
For the open-vocabulary 3D classification task, models are prompted using two query formats: (1) an instruction-style prompt—``\textit{What is this?}''—and (2) a completion-style prompt—``\textit{This is an object of ...}''.  
Our 7B and 13B models achieve evaluation scores of {57.11} and {58.48}, respectively, outperforming the best existing 3D MLLM (ShapeLLM) by +2.61 and +4.48.  
These results indicate that the proposed model facilitates stronger 3D–language alignment and generalization across prompts, enabling consistent recognition and reasoning about object semantics without the need for dedicated 3D encoders.

\textbf{3D Visual Question Answering.}
We further evaluate our model on the {MM-Vet} benchmark~\cite{shapellm} for 3D visual question answering (VQA). Our 7B model achieves a GPT-4 evaluation score of {49.53}, outperforming existing methods by 2.13. 
Scaling to 13B further boosts performance to {54.89}, marking a +1.79 improvement over ShapeLLM-13B. This consistent trend across all three tasks demonstrates that our encoder-free architecture effectively captures fine-grained geometric and semantic relationships, leading to more coherent and human-aligned multimodal reasoning.

\subsection{Analysis and Discussion}

\subsubsection{SAGE$^*$ Outperforms Existing 3D MLLM}
The primary goal of this work is to explore the feasibility of an end-to-end architecture for 3D MLLMs, aiming to alleviate the limitations inherent in pre-trained encoder-based designs. As shown in Table~\ref{tab:main}, our encoder-free model (SAGE$^*$)—trained under the same setup as prior works—surpasses or performs on par with existing methods that rely on pre-trained encoders. This demonstrates that the feature misalignment between pre-trained 3D encoders and the LLM input space is too large to be effectively bridged by a simple projection layer, as adopted in previous approaches. In contrast, our lightweight yet well-structured tokenizer learns a more coherent representation space, enabling the LLM to interpret 3D data analogously to a foreign language.

\begin{table}[t]
\centering
\caption{\textbf{Runtime Complexity Comparison.} 
Inference latency (ms) and memory (GB) on H100 GPU, 8K points, and Objaverse dataset.}
\label{tab:runtime}
\small
\begin{tabular}{lccc}
\toprule
\multirow{2}{*}{\textbf{Model}} & \textbf{Latency } & \textbf{Throughput } \\
& \textbf{ (ms) $\downarrow$} & \textbf{ (Sample/s) $\uparrow$} \\
\midrule
PointLLM-7B     & 239 & 4.2 \\
\rowcolor{linecolor4}\textbf{SAGE-7B} & \textbf{100} & \textbf{10.0}\\
\bottomrule
\end{tabular}
\vspace{-10pt}
\end{table}

\subsubsection{SAGE Is Computationally Efficient Than Encoder-based 3D MLLMs}

A key motivation behind the proposed encoder-free design of {SAGE} is to address the excessive latency and computational overhead introduced by traditional encoder-based 3D MLLMs. In existing methods such as PointLLM, the 3D encoder must first process dense point clouds into latent features before the language model can begin text generation. This additional preprocessing stage significantly slows inference, making real-time or large-scale deployment impractical.

To quantify these differences, we compare inference latency and throughput between {SAGE} and PointLLM on an H100 GPU using 8K-point inputs from the Objaverse dataset (Table~\ref{tab:runtime}). Our results show that {SAGE} achieves an inference latency of \textbf{100 ms}, representing more than a \textbf{2.3$\times$ speedup} over PointLLM (239 ms), while simultaneously improving throughput from 4.2 to \textbf{10.0 samples per second}. This efficiency gain stems from eliminating the heavy geometric encoder and replacing it with a lightweight geometric tokenizer that directly projects spatial features into the LLM’s embedding space.

\begin{figure}
    \centering
    \includegraphics[width=0.7\linewidth]{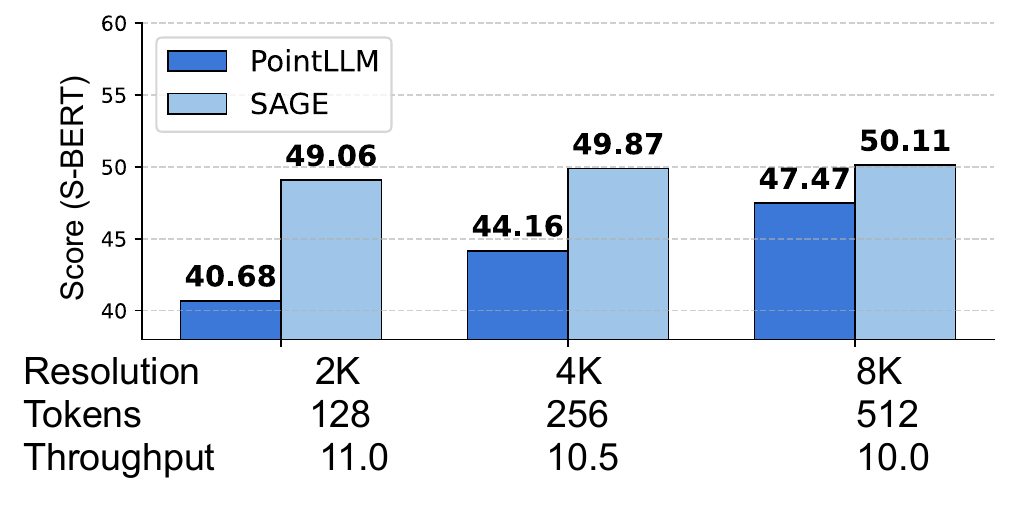}
    \caption{Performance of SAGE on diverse ranges of point cloud resolution on 3D captioning task on Objaverse dataset.}
    \label{fig:resolution_comparison}
    \vspace{-10pt}
\end{figure}

\begin{table*}[t]
\centering
\caption{\textbf{Qualitative results on Objaverse.} We adopt this table from \cite{pointllm}, and compare our proposed model's generated description for the specific 3D object samples. 
}

\label{tab:demo_objaverse}
\scalebox{0.98}{
\begin{tabular}{@{}l p{0.38\linewidth} p{0.38\linewidth}}
\toprule

Samples 1, 2 & 
  \begin{minipage}{\linewidth}
    \includegraphics[width=0.37\linewidth,height=0.26\linewidth]{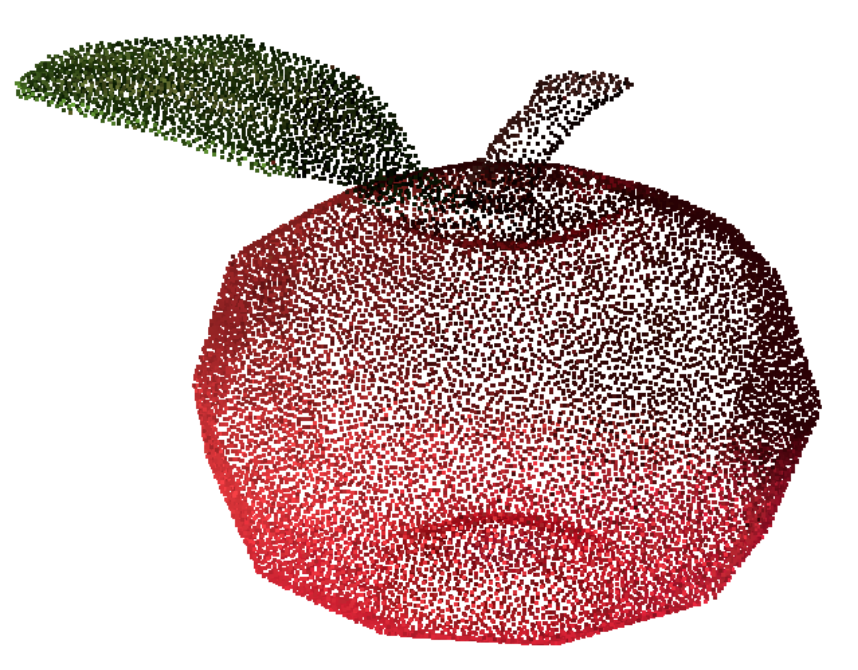}
  \end{minipage}
& 
  \begin{minipage}{\linewidth} 
    \includegraphics[width=0.39\linewidth,height=0.28\linewidth]{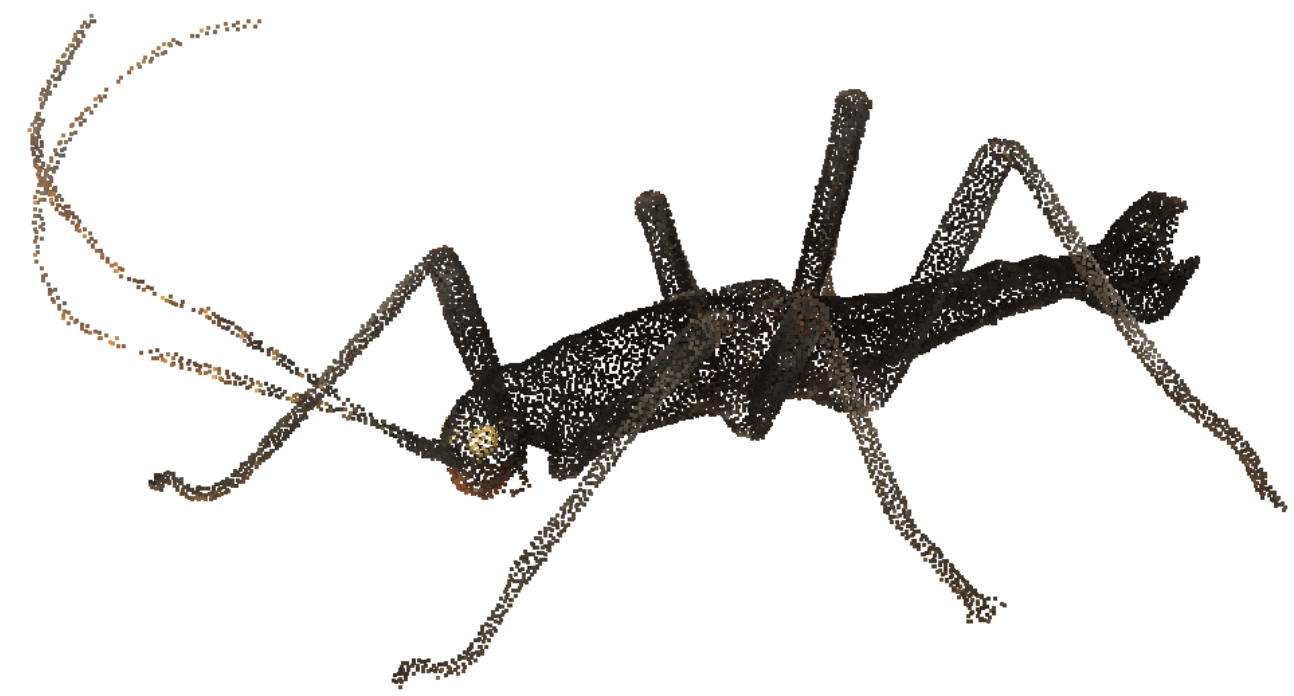}
  \end{minipage}
\\ \midrule
Uid & 0ea33b6617174530b97d6b7a92c275fb & de8ec2a724f14fc4b54624512f80f13e \\ \midrule

Prompt & What is this? & This is an object of  \\
Human & A cartoon green and red like a fruit & A black insect \\
InstructBLIP\cite{dai2023instructblip} & an appleavatar 3d model & 26 \\ 
LLaVA\cite{liu2023visual} & This is a computer-generated image (CGI) or a 3D model of an apple with a leaf on it. & This is an image of a large, close-up, and dark-colored insect, possibly a big cockroach, with long antennae. \\ 
3D-LLM\cite{3dllm} & A 3D model of a red apple. & A small, black spider with a long tail. \\
PointLLM \cite{pointllm} & This is a 3D model of a unique apple, distinctively adorned with a single, vibrant green leaf at the top. & This 3D model depicts a realistic, jet-black insect with a pair of striking, golden brown eyes. \\
\rowcolor{linecolor4}\textbf{SAGE}  & \textbf{This is a 3D model of an apple, which is vividly colored with a shiny red exterior and a contrasting green leaf at the top.} & \textbf{This is a 3D model of a type of insect, characterized by its six legs and two antennae, depicted in a distinctive black color.} \\
\bottomrule
\end{tabular}}
\vspace{-10pt}
\end{table*}

\subsubsection{Robustness to Varying Input Resolutions}

Another limitation of encoder-based models, such as PointLLM, lies in their fixed input resolution. When encountering high-resolution point clouds, these models must subsample the input, inevitably discarding fine-grained geometric details. Conversely, when presented with low-resolution inputs, they must upsample points to match the expected resolution—introducing redundant information that provides no additional semantic value while increasing inference overhead.
In contrast, SAGE naturally supports variable-resolution inputs, efficiently adapting to the density of the point cloud. It achieves higher efficiency with smaller inputs while maintaining strong performance on high-resolution data—offering the best of both worlds.

As shown in Figure~\ref{fig:resolution_comparison}, we evaluate performance across different input resolutions (2K, 4K, and 8K points) on the Objaverse dataset. Since PointLLM requires a fixed input resolution, its inputs are upsampled to the predefined resolution before processing. The results clearly show that PointLLM’s performance drops sharply on low-resolution data due to its inability to handle sparse inputs, whereas SAGE exhibits only a minor degradation, demonstrating its robustness to resolution variability. Additionally, the throughput of SAGE improves with lower resolution, as the number of input tokens drops. In contrast, PointLLM's throughput remains constant regardless of the original input resolution due to its fixed preprocessing requirement.

\subsection{Additional Analysis}
In the appendix, we present additional experiments, including comparisons to existing methods on different datasets (\textit{Appendix} \ref{app:compare}), performance comparison across multiple runs (\textit{Appendix} \ref{app:variance}) and a detailed ablation and sensitivity studies (\textit{Appendix} \ref{app:ablation_sensitivity}) on all the main components and hyperparameters.

\subsection{Qualitative Analysis}
Table \ref{tab:demo_objaverse} presents qualitative results on the Objaverse dataset, comparing the descriptive outputs of our proposed model, SAGE, against existing baselines. Each column shows examples of 3D object prompts alongside generated descriptions. As seen, SAGE produces more precise, contextually rich, and visually grounded descriptions, accurately capturing fine-grained geometric and appearance details (e.g., color, texture, and structure) that baseline models either overlook or describe vaguely. For example, for sample 2, SAGE can describe the number of different body parts of the sample. This highlights SAGE’s stronger alignment between 3D geometry and linguistic representation, leading to more natural and detailed 3D object understanding.

\section{Conclusion}
In this work, we presented SAGE, the first encoder-free 3D Multimodal Large Language Model that directly processes raw point clouds without relying on pre-trained 3D encoders. By introducing a lightweight 3D tokenizer that transforms geometric structures into discrete tokens through sampling and vector quantization, our approach effectively treats 3D data as a foreign language, enabling seamless integration within large language models. Furthermore, our multi-stage training pipeline, culminating in a reinforcement learning stage with semantic alignment–based rewards, enhances open-ended 3D reasoning capabilities where answers are descriptive rather than verifiable. Extensive experiments demonstrate that our encoder-free framework not only matches or surpasses the performance of encoder-based methods but also offers superior computational efficiency, backbone generalization, and robust resolution handling. We believe this work lays the foundation for future exploration of unified multimodal reasoning frameworks that treat diverse sensory modalities—such as 2D, 3D, and language—as components of a shared linguistic space.

\clearpage

{\small
\bibliographystyle{ieee_fullname}
\bibliography{ref}
}
\clearpage

\renewcommand{\thesection}{A\arabic{section}}
\renewcommand{\thefigure}{A\arabic{figure}}
\renewcommand{\thetable}{A\arabic{table}}

\setcounter{table}{0}
\setcounter{figure}{0}
\setcounter{section}{0}
\setcounter{page}{1}

\section*{Appendix}
In this section, we present more details on the implementation and discuss additional experimental results. The following is an overview of the organization of the appendix. \\

\begin{description}[noitemsep, topsep=0pt, parsep=0pt, partopsep=0pt]
    \item[\ref{app:implementation}] Implementation Details
    \item[\ref{app:discussion}] Discussion
    \begin{description}[noitemsep, topsep=0pt, parsep=0pt, partopsep=0pt]
        \item[\ref{app:discussion:don't-compare}] Non-comparable existing literature
    \end{description}
    \item[\ref{app:results}] Additional Results
    \begin{description}[noitemsep, topsep=0pt, parsep=0pt, partopsep=0pt]
        \item[\ref{app:compare}]  Comparison to Existing Methods
        \item [\ref{app:training}] Training Time Analysis
        \item [\ref{app:zero_shot}] Zero-shot Classification
        \item [\ref{app:variance}] Performance Across Multiple Runs
        
    \end{description}
    \item[\ref{app:ablation_sensitivity}] Ablation and Sensitivity Studies
    \begin{description}[noitemsep, topsep=0pt, parsep=0pt, partopsep=0pt]
        \item[\ref{app:sensitivity}]  Sensitivity Study
        \item [\ref{app:ablation}] Ablation Study
        \item [\ref{app:generalization}] Generalization Across LLM Backbones
    \end{description}
    
    \item[\ref{app:qualitative}] More Qualitative Results

    \end{description}

\section{Implementation Details}\label{app:implementation}
We discussed most of the training and implementation details in the main paper. In this section, we summarize all hyperparameters in Table \ref{tab:hyperparams_list}. Some parameters are shared across all training stages, while others are stage-specific. The values of these hyperparameters are chosen based on a rigorous sensitivity analysis, while some are derived from well-established literature. 

Next, we present the prompt used for GPT-4 evaluation in Prompt 1. This prompt is similar to what existing methods have used for this evaluation. 

\begin{table}[h]
\centering
\caption{List of model-specific hyperparameters.}
\label{tab:hyperparams_list}
\small
\begin{tabular}{llc}
\toprule
\textbf{Stage} & \textbf{Hyperparam.} & \textbf{Value} \\
\midrule
    &  $\beta$ & 0.25 \\
  &  $\lambda$ & 0.50 \\
  & Codebook size & 8192 \\
 & Warm-up ratio   & 0.03 \\
 & Weight decay & 0.05 \\ 
 
  & Number of sampled points, $N_s$ & 512 \\
  & Number of neighbours, $K_g$ & 81 \\
  \hline
   
 1 & Epoch    & 3 \\
 & Lr. rate    & $4\times10^{-4}$ \\
  & Batch size & 128\\ \hline
 2 & Epoch  & 3 \\
 & batch size & 32 \\
 & Lr. rate & $2\times10^{-5}$ \\ \hline
3 &  $\alpha$ & 0.95 \\
 & Epoch  & 1 \\
    & batch size & 8 \\
 & Lr. rate & $1\times10^{-6}$ \\
 
 & number of generated responses, $m$ & 8 \\
\bottomrule
\end{tabular}
\end{table}

\noindent\fbox{\begin{minipage}{\columnwidth}
{\footnotesize
\noindent \texttt{
\textbf{Prompt 1:} \\
You are a helpful AI assistant. \\
Now, I will give you a question, its type, an answer from the model, and an answer from the label. \\
Your task is to focus only on these two answers and determine whether they convey the same information for the given type of question. \\
Your response should be a single confidence score ranging from 0 to 100. \\
This score evaluates how closely the two answers describe the same thing. \\
Follow the scoring standard demonstrated below. \\
Here are several example question-answer pairs with their confidence scores: \\
\\
Question 1: How many oranges will there be if 1/3 of them are removed?\\
Question type: Knowledge\\
Answer from model: There will be 6 left.\\
Answer from label: As there are 9 oranges in total, there will be 6 oranges left if 1/3 of them are removed.\\
Confidence score: 100\\
\\
Question 2: What is this object?\\
Question type: General Visual Recognition\\
Answer from model: This is a bathtub.\\
Answer from label: This is a dirty bathtub.\\
Confidence score: 80\\
\\
Question 3: What is this object?\\
Question type: General Visual Recognition\\
Answer from model: This is a bottle of water.\\
Answer from label: This is a bottle of oil.\\
Confidence score: 50\\
\\
Question 4: What is the boy holding in his right hand?\\
Question type: Spatial Recognition\\
Answer from model: He is holding a white cup in his right hand.\\
Answer from label: He is holding a sword in his right hand.\\
Confidence score: 0\\
\\
Next, you will be given: \\
Question: \{\},\\
Question type: \{\},\\
Answer from model: \{\},\\
Answer from label: \{\}.\\
Output only the confidence score as a number, without any words.
}
}
\end{minipage}}

\begin{table*}
    \caption{
        \textbf{Performance comparison with existing methods.} We adopt the following table from \cite{3dllava} and follow the notations and categorization defined by them. Here,
        “Specialist Model” refers to models specifically designed for individual tasks such as 3D question answering, 3D dense captioning, or referring segmentation.
        “Finetuned 3D MLLMs” denotes models that are jointly trained and subsequently fine-tuned on each dataset before evaluation.
        “3D MLLMs” represents models trained on multiple tasks without task-specific fine-tuning.
        “PC” stands for point cloud, and “I” denotes multi-view images.
        Note that the results of LEO~\cite{huang2023embodied} on ScanQA are shown in gray and excluded from direct comparison, as the model uses a different setting that accesses ground-truth objects related to the questions.
        All models use the 7B parameter configuration.
    }
    \renewcommand{\arraystretch}{1.05}
    \centering
    \resizebox{0.65\linewidth}{!}{
    \begin{tabular}{l|c|ccc|ccc}
    \toprule
     & & \multicolumn{3}{c|}{\textbf{ScanQA (val)}} & \multicolumn{3}{c}{\textbf{Scan2Cap (val)}} \\
      \textbf{Method} & \textbf{Modality} & \textbf{B-4}$\uparrow$ & \textbf{M}$\uparrow$ & \textbf{R}$\uparrow$ & \textbf{B-4}$\uparrow$ & \textbf{M}$\uparrow$ & \textbf{R}$\uparrow$ \\ 
      \midrule[0.6pt]
    \multicolumn{8}{c}{\textit{Specialist Models}}\\
    \midrule[0.6pt]
    ScanQA\cite{azuma2022scanqa} & PC & 10.1 & 13.1 & 33.3 & - & - & - \\
    3D-VLP\cite{jin2023context} & PC & 11.2 & 13.5 & 34.5 & 32.3 & 24.8 & 51.5 \\
    3D-VisTA\cite{zhu20233d} & PC & 10.4 & 13.9 & {\textbf{45.7}} & 34.1 & 26.8 & 55.0 \\
    Scan2Cap\cite{chen2021scan2cap} & PC & - & - & - & 23.3 & 22.0 & 44.8 \\
    MORE\cite{jiao2022more} & PC & - & - & - & 22.9 & 21.7 & 44.4 \\
    SpaCap3D\cite{wang2022spatiality} & PC & - & - & - & 25.3 & 22.3 & 45.4 \\
    D3Net\cite{chen2021d3net} & PC & - & - & - & 30.3 & 24.4 & 51.7 \\
    UniT3D\cite{chen2023unit3d} & PC & - & - & - & 27.2 & 21.9 & 46.0 \\
    3DJCG\cite{cai20223djcg} & PC & - & - & - & 31.0 & 24.2 & 50.8 \\
    Vote2Cap-DETR~\cite{chen2023end} & PC & - & - & - & 34.5 & 26.2 & 54.4 \\
    \midrule[0.6pt]
    \multicolumn{8}{c}{\textit{Finetuned 3D MLLMs}}\\
    \midrule[0.6pt]
    3D-LLM\cite{3dllm} & PC+I & 12.0 & 14.5 & 35.7 & - & - & - \\
    Scene-LLM~\cite{fu2024scene} & PC+I & 12.0 & 16.8 & 40.0 & - & - & - \\
    LL3DA ~\cite{ll3d} & PC & 13.5 & 15.9 & 37.3 & 36.8 & 26.0 & 55.1 \\
    \midrule[0.6pt]
    \multicolumn{8}{c}{\textit{3D MLLMs}}\\
    \midrule[0.6pt]
    LEO~\cite{huang2023embodied} & PC+I & \textcolor{gray}{13.2} & \textcolor{gray}{20.0} & \textcolor{gray}{49.2} & {\textbf{38.2}} & 27.9 & \underline{{58.1}} \\
    Scene-LLM~\cite{fu2024scene} & PC+I & 11.7 & 15.8 & 35.9 & - & - & - \\
    Chat-Scene~\cite{huang2024chat} & PC+I & 14.3 & 18.0 & 41.6 & 36.4 & \underline{{28.0}} & \underline{{58.1}} \\
    Grounded 3D-LLM~\cite{chen2024grounded} & PC & 13.4 & - & - & 35.5 & - & - \\
    3D-LLaVA \cite{llava3d} & PC & \underline{17.1} & \underline{18.4} & 43.1 & {36.9} & 27.1 & 57.7 \\
    \rowcolor{linecolor4} \textbf{SAGE-7B} & PC & \textbf{17.5} &  \textbf{19.6}  & \underline{44.5} & \underline{37.8} & \textbf{28.1} & \textbf{58.6} \\
    \bottomrule
    \end{tabular}
    }
    \label{tab:additional}
    \vspace{-1pt}
\end{table*}

\section{Discussion}\label{app:discussion}

\subsection{Non-comparable Existing Method}\label{app:discussion:don't-compare}
While we discuss many existing methods in the related work, we do not directly compare our approach with all of them in the main results. This is because some of the existing methods differ substantially in terms of both their learning strategies and the types of data they use. For instance, some methods are first trained on dense prediction tasks \cite{3dllava} using segmentation data. Another major source of disparity lies in the training data. While our method, and a few others \cite{pointllm} adhere strictly to single training dataset \cite{deitke2023objaverse}, some approaches \cite{shapellm} combine multiple datasets from different sources \cite{collins2022abo, chang2015shapenet, fu20213d, deitke2023objaverse, modelnet, scanobject} to increase data volume, and others propose data generation pipelines to dynamically curate additional samples \cite{robin3d}. Additionally, prior works \cite{zhang2022pointclip, zhu2023pointclip, qi2023contrast, xue2023ulip, liu2023openshape, zhang2024tamm, zhou2023uni3d, gao2024sculpting, xue2024ulip} in this literature focused on 3D encoders with any multimodal capabilities utilizing well-established training pipelines \cite{ilharco2021openclip, sun2023eva}. In the main table of the paper, we only compare against methods that follow similar datasets and training protocols \cite{pointllm} as ours. Nonetheless, we provide additional results and further discussion on these disparities in the appendix.

\begin{figure*}[t]
\centering
\begin{minipage}[b]{0.22\textwidth}
  \centering
  \includegraphics[width=\linewidth]{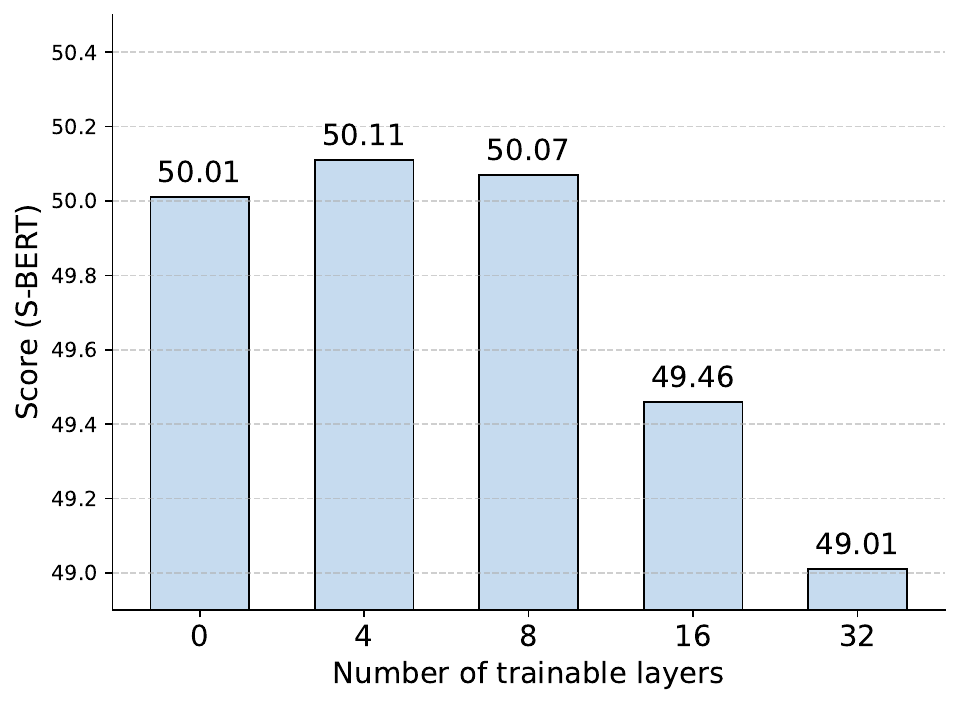}
  \vspace{-5pt}
  \label{fig:resolution}
\end{minipage}
~~~~~~~~~~~~~~~~
\begin{minipage}[b]{0.22\textwidth}
  \centering
  \includegraphics[width=\linewidth]{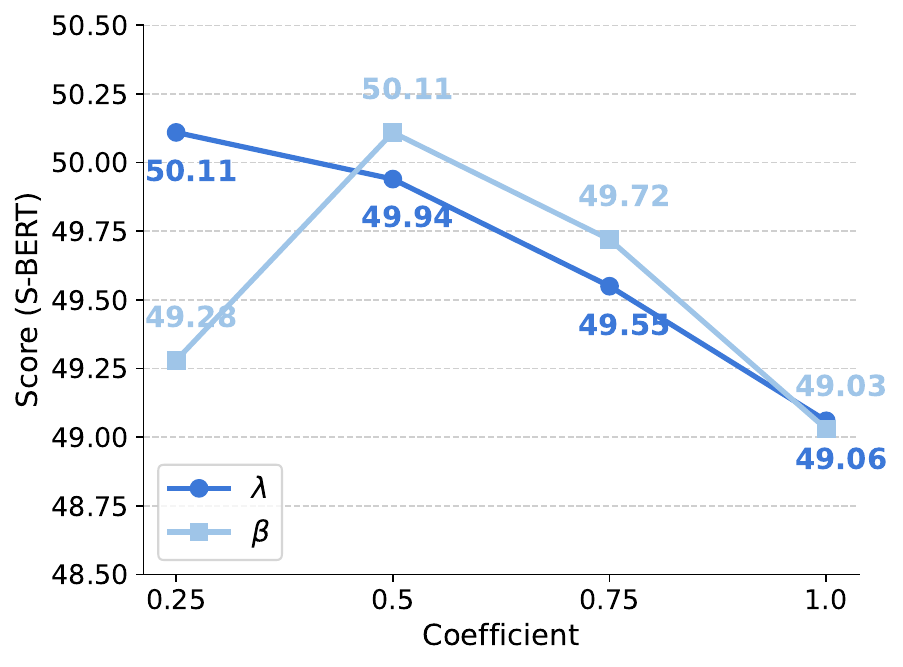}
  \vspace{-1mm}
  \label{fig:llm_trainable_layer}
\end{minipage}
~~~~~~~~~~~~~~~~
\begin{minipage}[b]{0.22\textwidth}
  \centering
  \includegraphics[width=\linewidth]{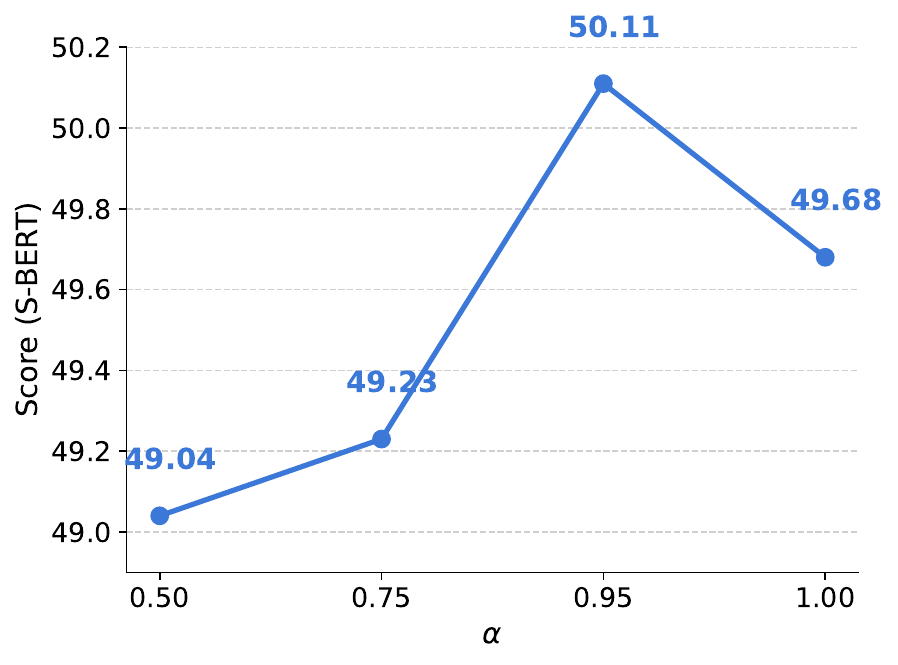}
  \vspace{-1mm}
  \label{fig:llm_trainable_layer}
\end{minipage}
\vspace*{-5pt}
\caption{Fig. (Left) Impact of the number of LLM trainable layers during the stage
1 training. Fig. (Middle) Impact of the number of LLM trainable layers during stage 1 training. Fig(Right) Impact of group normalization coefficient on performance of SAGE-7B.}
\vspace{-10pt}
\label{fig:performance}
\end{figure*}

\begin{table}
    
    \small
    \setlength    \tabcolsep{5pt}
    \caption{Total training time.}
    \vspace{-10pt}
    \begin{center}
    \vspace{-10pt}
    \begin{tabular}{lc }
    \midrule
    \textbf{Model} & \textbf{Training time (h)}\\ \hline
     PointLLM & 26.1\\
     SAGE$^*$ & 18.0\\
     SAGE & 27.4\\
    \bottomrule
    \end{tabular}
    \vspace{-20pt}
    \label{tab:training_time}
    \end{center}
\end{table}

\section{Additional Results}\label{app:results}
In this section, we present additional experimental results for our proposed model. We first compare SAGE’s performance with a broader set of state-of-the-art methods on 3D captioning and 3D VQA tasks across two datasets. We then report multi-run results on the Objaverse 3D captioning task to demonstrate the stability of both the model and the training pipeline. Finally, we include a training-time analysis comparing SAGE with existing approaches in the literature.

\subsection{Comparison to Existing Methods}\label{app:compare}
In Table \ref{tab:additional}, we compare SAGE with additional existing SOTA methods on 3D VQA and captioning tasks using the ScanQA \cite{azuma2022scanqa} and Scan2Cap \cite{chen2021scan2cap} datasets, respectively. On both 3D VQA and captioning tasks, SAGE outperforms the existing SOTA methods on most metrics. While no existing method dominantly performs best across all metrics, SAGE surpasses all existing methods in 4 out of 6 metrics across the 2 datasets, despite using less data for training and being computationally more efficient than existing methods.

\subsection{Training Time Analysis }\label{app:training}
In this section, we compare the total training time of our proposed method with PointLLM, which uses the same amount of training data and a similar training setup (Table \ref{tab:training_time}). Here, the total training time refers to the end-to-end duration of the entire training pipeline. Due to its efficient model design, SAGE$^*$ requires considerably less training time (18 hours compared to 26.1 hours for PointLLM). The preference optimization step introduces additional training time; however, the complete three-stage training of our method still takes only slightly longer than the two-stage training pipeline of PointLLM.

\subsection{Zero-shot Classification} \label{app:zero_shot}
In Table \ref{tab:zeroshot}, we compare the zero-shot classification performance of SAGE with existing SOTA methods on the ModelNet40 dataset. Here, SAGE shows improved or comparable performance with existing SOTA methods.

\begin{table}[t!]
\scriptsize
  \setlength{\tabcolsep}{4pt}
  \centering
  \caption{\textbf{Performance comparison on Zero-shot 3D classification} on ModelNet40~\cite{modelnet} and ScanObjectNN~\cite{scanobject}. Here, Ensembled~\cite{liu2023openshape} reprsents pretraining with four datasets, ShapeNet~\cite{chang2015shapenet}, ABO~\cite{collins2022abo} and 3D-FUTURE~\cite{fu20213d}. $^\dagger$: Uni3D employs a larger EVA-CLIP-E~\cite{sun2023eva} teacher, while other methods employ OpenCLIP-bigG~\cite{ilharco2021openclip}.} Ours uses the Objaverse dataset paired with Cap3D generated captions.
  \vspace{-8pt}
  \resizebox{0.86\linewidth}{!}{
    \begin{tabular}{lcccccc}
    \toprule[0.95pt]
    \multirow{2}{*}[-0.5ex]{Method} & \multicolumn{3}{c}{\textcolor{black}{\textbf{ModelNet40}}} & \multicolumn{3}{c}{\textcolor{black}{\textbf{ScanObjectNN}}}\\
    \cmidrule{2-7} & Top1 & Top3 & Top5 & Top1 & Top3 & Top5 \\
    \midrule[0.6pt]
    \multicolumn{7}{c}{\textit{2D Inference without 3D Training}}\\
    \midrule[0.6pt]
    PointCLIP~\cite{zhang2022pointclip} & 19.3 & 28.6 & 34.8 & 10.5 & 20.8 & 30.6\\
    PointCLIPv2~\cite{zhu2023pointclip} & 63.6 & 77.9 & 85.0 & 42.2 & 63.3 & 74.5\\
    \midrule[0.6pt]
    \multicolumn{7}{c}{\textit{Trained on ShapeNet}}\\
    \midrule[0.6pt]
    {\scshape ReCon}~\cite{qi2023contrast} & 61.2 & 73.9 & 78.1 & 42.3 & 62.5 & 75.6\\
    CLIP2Point~\cite{CLIP2Point} & 49.5 & 71.3 & 81.2 & 25.5 & 44.6 & 59.4\\
    ULIP~\cite{xue2023ulip} & 60.4 & 79.0 & 84.4 & 51.5 & 71.1 & 80.2\\
    OpenShape~\cite{liu2023openshape} & 70.3 & 86.9 & 91.3 & 47.2 & 72.4 & 84.7\\
    TAMM~\cite{zhang2024tamm} & 73.1 & 88.5 & 91.9 & 54.8 & 74.5 & 83.3\\
    MixCon3D~\cite{gao2024sculpting} & 72.6 & 87.1 & 91.3 & 52.6 & 69.9 & 78.7\\
    \midrule[0.6pt]
    \multicolumn{7}{c}{\textit{Trained on Ensembled}}\\
    \midrule[0.6pt]
    ULIP-2~\cite{xue2024ulip} & 75.1 & 88.1 & 93.2 & 51.6 & 72.5 & 82.3\\
    OpenShape~\cite{liu2023openshape} & 84.4 & 96.5 & 98.0 & 52.2 & 79.7 & 88.7\\
    MixCon3D~\cite{gao2024sculpting} & {86.8} & \textbf{96.9} & \textbf{98.3} & 58.6 & 80.3 & 89.2\\
    Uni3D-B$^\dagger$~\cite{zhou2023uni3d} & 86.3 & 96.5 & 97.9 & {63.8} & {82.7} & 90.2\\
    Uni3D-L$^\dagger$~\cite{zhou2023uni3d} & 86.3 & {96.8} & \textbf{98.3} & 58.2 & 81.8 & 89.4\\
    {\scshape ReCon++}-B \cite{shapellm} & 86.5 & 94.7 & 95.8 & 63.6 & \textbf{84.2} & \textbf{90.6} \\
    \rowcolor{linecolor4} \textbf{SAGE-7B}  & \textbf{88.9} & 94.7 & \textbf{98.3} & \textbf{65.8} & 80.2 & \textbf{90.6} \\
    \bottomrule[0.95pt]
    \end{tabular}
    }
  \label{tab:zeroshot}
\end{table}

\begin{table*}
\centering
\caption{Performance comparison across \textit{three} runs on \textit{(SAGE)} and \textit{(SAGE$^*$)} on 7B parameters.  Here, the Objaverse dataset is used for 3D object captioning and recognition tasks, and the MM-Vet dataset is used for 3D VQA.}
\label{tab:variance}
\small
\begin{tabular}{l|cccccc|c|c}
\hline
\multirow{2}{*}{\textbf{Model}} & \multicolumn{6}{c|}{\textbf{Captioning}} & {\textbf{Cls.}} & {\textbf{VQA}} \\
\cline{2-9}
& \textbf{GPT-4} & \textbf{Sentence-BERT} & \textbf{SimCSE} & \textbf{BLEU-1} & \textbf{ROUGE-L} & \textbf{METEOR} & \textbf{GPT-4} & \textbf{GPT-4} \\
\hline
SAGE-7B$^*$ & 49.05 & 49.23 & 48.56 & 7.41 & 10.25 & 14.35 & 55.71 &  46.38\\
\textit{SD} & 0.09 & 0.04 & 0.05 & 0.04 & 0.02 & 0.02 & 0.03 &  0.04\\ \hline
SAGE-7B & {50.98} & 50.11 & 49.70 & 9.50 & 12.66 & 16.95 & 57.11 &  49.53\\
\textit{SD} & 0.08 & 0.03 & 0.04 & 0.05 & 0.04 & 0.03 & 0.03 &  0.05\\ 
\hline
\end{tabular}
\vspace{-10pt}
\end{table*}

\begin{table*}[t]
\centering  
\caption{Sensitivity study on different model-specific parameters on ModelNet40.}
\vspace{-5pt}
\resizebox{0.75\linewidth}{!}{
    \subfloat[ Codebook size. \vspace{5pt} \label{tab:codebook_size}]{ 
        \begin{minipage}{0.2\linewidth}
            \centering  
            \small
            \begin{tabular}{l c}
                \toprule
                \textbf{Size} & \textbf{S-BERT $\uparrow$} \\
                \midrule
                4096    & 48.88 \\
                8192    & 50.11 \\
                16384  & 49.76 \\
                \bottomrule
            \end{tabular}
        \end{minipage}
    }
    \hspace{2em}
    \subfloat[Number of point tokens. \vspace{5pt} \label{tab:pc_token}]{ 
        \begin{minipage}{0.2\linewidth}
            \centering
            \small
            \begin{tabular}{l c}
                \toprule
                \textbf{Size} & \textbf{S-BERT $\uparrow$} \\
                \midrule
                128    & 48.98 \\
                512    & 50.11 \\
                1024  & 50.09 \\
                \bottomrule
            \end{tabular}
        \end{minipage}
    }
    \hspace{2em}
    \subfloat[Types of pooling layers. \vspace{5pt} \label{tab:pooling}]{ 
        \begin{minipage}{0.2\linewidth}
            \centering
            \small
            \begin{tabular}{l c}
                \toprule
                \textbf{Size} & \textbf{S-BERT $\uparrow$} \\
                \midrule
                Max pool    & 50.11 \\
                Avg. pool    & 49.89 \\
                Attn. pool  & 48.03 \\
                \bottomrule
            \end{tabular}
        \end{minipage}
    }
}
\vspace{-5pt}
\label{tab:ablation_ema}
\end{table*}

\begin{table}
    \vspace{-5pt}
    \small
    \setlength    \tabcolsep{5pt}
    \caption{Impact of discrete vs. continuous point embeddings.}
    \vspace{-10pt}
    \begin{center}
    \begin{tabular}{cc}
    \hline
    \textbf{Embedding type} & \textbf{S-BERT $\uparrow$}\\ \hline
    Continuous (w/o codebook) & 47.67\\
    Discrete (with codebook) & 50.11\\ 
    \hline
    \end{tabular}
    \label{tab:abl_codebook}
    \end{center}

    \vspace{-20pt}
\end{table}

\begin{table}
    \vspace{-5pt}
    \small
    \setlength    \tabcolsep{5pt}
    \caption{Impact of different LLM backbones.}
    \vspace{-10pt}
    \begin{center}
    \begin{tabular}{ccc }
    \hline
    \textbf{LLM} & \textbf{Model} & \textbf{S-BERT $\uparrow$}\\ \hline
    LLaMA-3.1 & PointLLM & 51.23\\
     & SAGE & \textbf{55.89}\\ \hline
     Qwen-2.5 & PointLLM & 52.35\\
     & SAGE & \textbf{56.91}\\
    \hline
    \end{tabular}
    \label{tab:abl_llm}
    \end{center}
    \vspace{-20pt}
\end{table}

\subsection{Performance Across Multiple Runs}\label{app:variance}
To assess the stability of our method, we report the mean and standard deviation of SAGE and SAGE$^*$ (7B parameters) across three independent runs in Table \ref{tab:variance}. Both variants demonstrate consistent performance with low variance across all metrics.

\section{Ablation and Sensitivity Studies}\label{app:ablation_sensitivity}
In this section, we perform an in-depth sensitivity analysis of various model-specific hyperparameters. We also conduct a detailed ablation study on the key components of our proposed model. All experiments are carried out using the 7B-parameter setting on the 3D captioning task with the Objaverse dataset, and we report the S-BERT alignment score.

\subsection{Sensitivity Study}\label{app:sensitivity}

\textbf{\begin{figure*}[t!]
\centering
\includegraphics[width=0.96\linewidth]{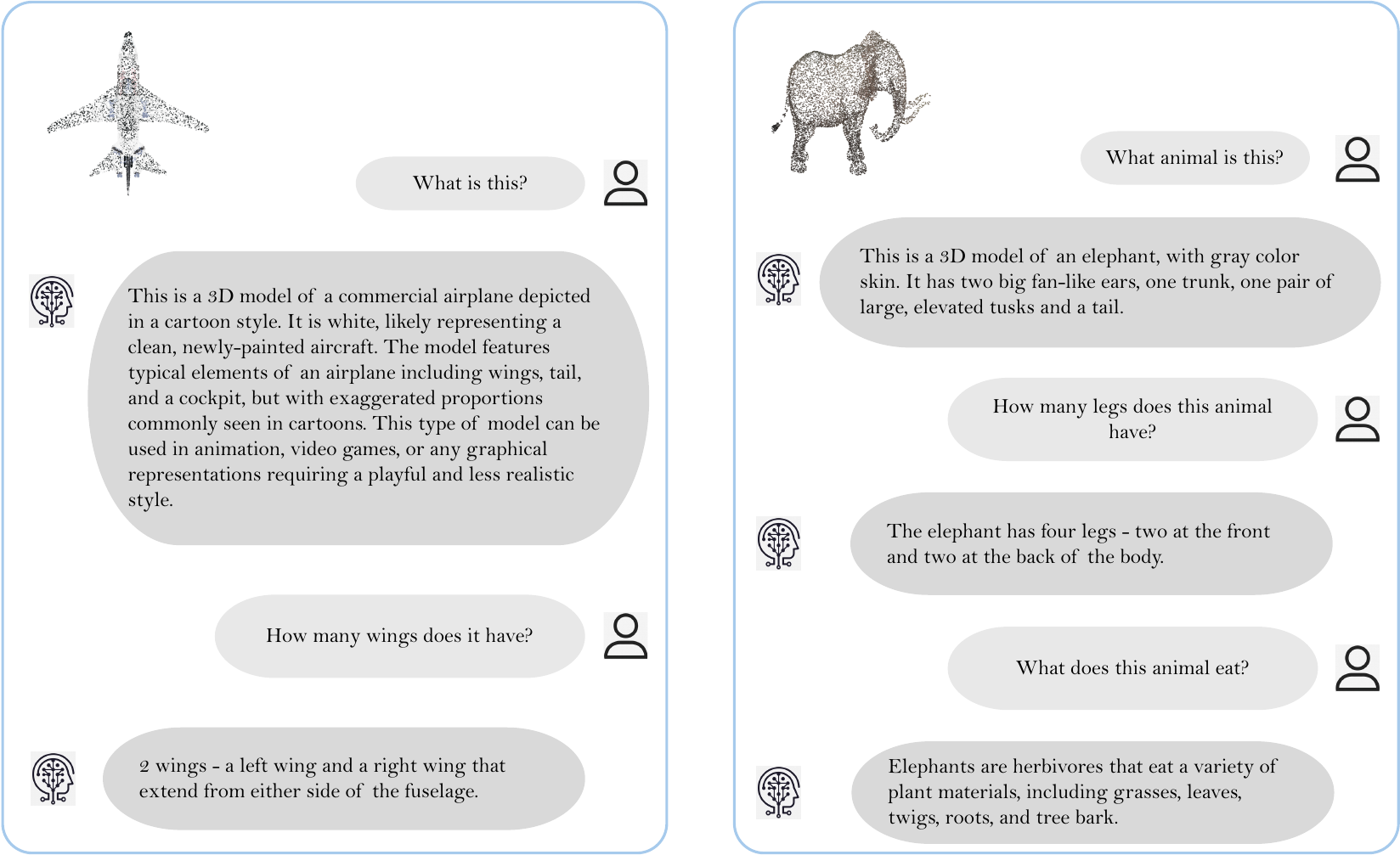}
\captionof{figure}{Dialogues between PointLLM and a human user.}
\label{fig:dialogue}
\end{figure*}}

\begin{table*}[t]
\centering
\caption{Qualitative results on the Objaverse dataset.}
\label{tab:demo_objaverse}
\scalebox{0.98}{
\begin{tabular}{@{}l p{0.38\linewidth} p{0.38\linewidth}}
\toprule
Samples 1, 2 &
  \begin{minipage}{\linewidth}
    \includegraphics[width=0.37\linewidth,height=0.30\linewidth]{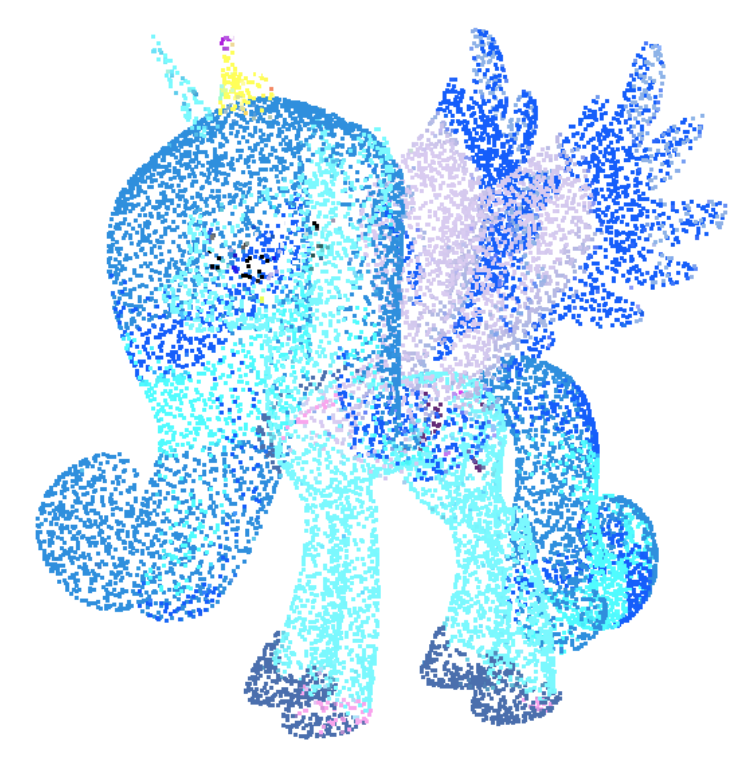}
  \end{minipage}
&
  \begin{minipage}{\linewidth}
    \includegraphics[width=0.40\linewidth,height=0.18\linewidth]{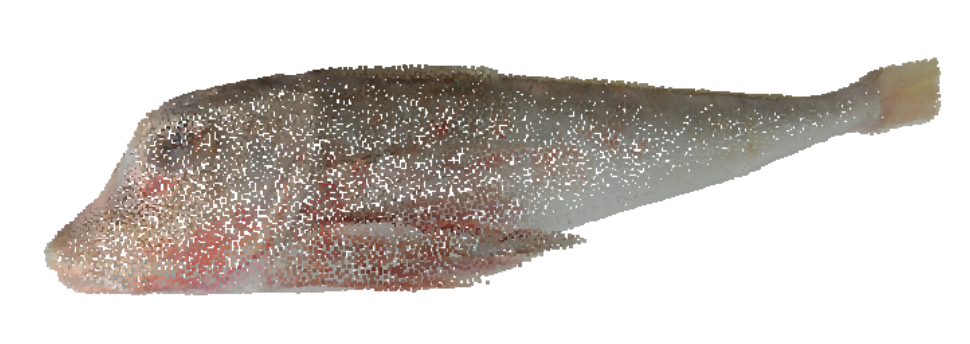}
  \end{minipage}
\\ \midrule
Uid & 059a7936ed89419ba9eae3153753ae86 & 82413f74efb84bc3b872a7a8f26cca81 \\ \midrule
Prompt & Caption this 3D model in detail. & Caption this 3D model in detail. \\
Human & A blue cartoon & A white fish with orange/red spots \\
InstructBLIP\cite{dai2023instructblip} & There is a cute blue and white pony in a still, looking cheerful and cheerful. It's standing next to a larger blue-colored object, potentially a barrier or structure, which provides the backdrop for this scene. The overall image is designed to create a fun, whimsical setting for the pony. & There is a fish hanging on a string in front of a black background. The fish is swimming in the middle of the image and is a little bit up from the bottom of the screen. The fish has a red spots on the left and right sides, and is positioned slightly towards the top of the screen. \\
LLaVA\cite{liu2023visual} & The image features a 3D model of a cute, colorful cartoon unicorn with a blue and white body and a hint of pink. The unicorn has a magical aura and is set against a black background. Its long, flowing mane and tail add to its whimsical charm. The unicorn appears to be in motion, possibly flying or prancing, as it captures the viewer's imagination. & The image features a 3D model of a fish floating in a dark, empty space. The fish is positioned towards the left side of the image and appears to be swimming upwards. The fish is large and occupies a significant portion of the image, giving the impression of a life-like, computer-generated representation. The dark background and lack of other objects in the scene emphasize the fish's presence and create a sense of depth and dimension. \\
3D-LLM\cite{3dllm} & A 3D model of a purple and pink pony with long hair, standing on its hind legs. & A small fish with a long tail is visible in the image. It has no distinguishing features or markings, and its color appears to be brown. \\
PointLLM\cite{pointllm} & A 3D model of a vibrant blue cartoon pony, characterized by its large white eyes, two wings, and a distinctive horn. The creature's overall design exudes a playful and friendly aura, making it appropriate for children's animations or toy production. The noticeably large white eyes and the unique wing and horn features endow the pony with an element of fantasy, while its blue body color gives it an overall appealing look. & The 3D model is of a long fish, illustrating a striking color combination of red, yellow, and grey. It's characterized by a slender tail. The yellow color is mostly seen on the body of the fish, fading into a subtler, grayish hue towards the fins and tail. The model depicts a typical long fish species, often found in the ocean, and can be used in a variety of digital applications, like animation, games, or virtual reality. \\
\rowcolor{linecolor4}\textbf{SAGE} & \textbf{The 3D model is a depiction of a light blue pony, characterized by its vibrant color, large white eyes, and a notably long tail. It also has two vibrantly colored wings and a horn, which makes it look like a cartoon avatar.} & \textbf{The 3D model represents a fish with a streamlined body with a vibrant mix of color of yellow, red and silver. It has a long dorsal fin, two pectoral fins, a long anal fin, and large, captivating eyes.} \\
\bottomrule
\end{tabular}
}
\end{table*}

\textbf{Trainable layers.} Our model, SAGE, utilizes LLaMA as the pretrained LLM backbone, which consists of 32 layers. In our proposed training pipeline, during the 3D tokenizer warm-up stage (stage 1), we train the first 4 layers of the LLM backbone. Since our proposed method doesn't utilize any pretrained 3D encoder, here the LLM solely extracts rich features from the point tokens along with the text tokens. In this experiment, we evaluate several configurations of the LLM backbone, ranging from fully freezing the LLM to partially tuning 4, 8, or 16 layers, as well as full fine-tuning of all 32 layers. From Figure \ref{fig:performance}\textit{(left)}, we observe that the model's performance improves as 4 layers are trained compared to fully frozen LLM. Beyond this point, increasing the number of trainable layers leads to a performance drop, due to overfitting, given the relatively small size of the training dataset.

\textbf{Model hyperparameters.}
In the vector quantization loss, $\beta$ is a crucial hyperparameter that controls the contribution of the codebook loss and the quantization loss. Another important hyperparameter is $\lambda$, which balances the two components of the total training objective: the next-token prediction loss and the vector quantization loss. We investigate the impact of these two coefficients on SAGE's performance through a sensitivity study. As shown in Figure \ref{fig:performance} (middle), SAGE-7B achieves the highest performance gain with a quantization coefficient of 0.5 and a regularization coefficient of 0.25. Moreover, we observe that the performance is not highly sensitive to the choice of these hyperparameters.

Similarly, we examine the effect of the coefficient $\alpha$ in the reward combination during the policy optimization stage (stage 3). As shown in Figure \ref{fig:performance} (right), the performance of SAGE-7B improves as $\alpha$ increases, reaching its highest point at $\alpha = 0.95$, which indicates that performance benefits from a lower weight on the length reward. However, performance drops when this reward is completely removed ($\alpha = 1.0$), highlighting the importance of using both rewards.

\textbf{Codebook size.}
During vector quantization, the size of the learnable codebook defines the number of unique tokens allocated to the point cloud representations.
Table \ref{tab:codebook_size} presents a sensitivity study on the codebook size used in the model. We vary the codebook size across three settings: 4096, 8192, and 16384. We observe that increasing the codebook size from 4096 to 8192 improves the score by 1.23. However, further increasing it to 16384 leads to a performance drop, indicating that around 8k+ tokens can sufficiently represent the point cloud features when treating it as a foreign language to LLM.

\textbf{Number of point tokens.} 
The number of point tokens defines the number of discrete geometric units the model uses to represent a 3D input, directly controlling the level of detail and the computational cost in an MLLM. Table \ref{tab:pc_token} presents a sensitivity study on the number of point tokens used in the model. We vary this number across three settings: 128, 512, and 1024. Increasing the number of point tokens from 128 to 512 improves the score from 48.98 to 50.11. However, further increments do not lead to an improvement, even though they increase the computation. This suggests that 512-point tokens strike the optimal balance, maximizing the model’s alignment capability. Therefore, we stick to 512 tokens for our final model. 

\textbf{Types of pooling functions.}
In table \ref{tab:pooling}, we investigate different pooling strategies for aggregating token representations in the proposed model. We evaluate three variants: max pooling, average pooling, and attention-based pooling. Here, using the max pooling gives the highest performance of 50.11, while average and attention pooling yield 49.89 and 49.03, respectively. These results indicate that max pooling is the most effective aggregation mechanism, due to its ability to preserve salient feature signals critical for cross-modal alignment.

\subsection{Ablation Study}\label{app:ablation}

\textbf{Comparison of discrete vs. continuous embeddings.}\label{app:descrete_continuous}
 As discussed earlier, vector quantization bridges the gap between continuous geometric features and discrete language tokens by using a learnable codebook that maps continuous embeddings into a finite vocabulary of 3D tokens, effectively extending the LLM tokenizer to the 3D domain. To further validate its importance, we conduct an ablation study by comparing performance with and without the codebook in the 3D tokenizer. In Table \ref{tab:abl_codebook}, we can observe that with the codebook---using discrete embeddings---SAGE achieves a 50.11 score. Removing the codebook and keeping the embeddings continuous leads to a performance drop of 2.44. This confirms the necessity of vector quantization for enabling the LLM to effectively learn from multimodal signals.

\subsection{Generalization Across LLM Backbones} \label{app:generalization}
Finally, we investigate the generalization of SAGE to newer LLM backbones, namely LLaMA-3.1-8B and Qwen-2.5-7B. We also reproduce the results of PointLLM using these backbones. As shown in Table \ref{tab:abl_llm}, SAGE demonstrates strong performance on both newer LLMs without any additional parameter tuning, and it consistently outperforms PointLLM.

\section{More Qualitative Results}
\label{app:qualitative}
We present more qualitative results on Objaverse datasets in Table \ref{tab:demo_objaverse}. We further show interactive dialogues between a user and SAGE-7B in Figure \ref{fig:dialogue}, highlighting the model’s strong understanding of point-cloud geometry, appearance, and functional attributes. The examples further demonstrate SAGE’s ability to respond to user instructions with appropriate reasoning while avoiding biased or ill-informed outputs.



\end{document}